# BUILDING HIGH-ACCURACY MULTILINGUAL ASR WITH GATED LANGUAGE EXPERTS AND CURRICULUM TRAINING

*Eric Sun, Jinyu Li, Yuxuan Hu, Yimeng Zhu, Long Zhou, Jian Xue, Peidong Wang, Linquan Liu Shujie Liu, Edward Lin, Yifan Gong*

Microsoft

## ABSTRACT

We propose gated language experts and curriculum training to enhance multilingual transformer transducer models without requiring language identification (LID) input from users during inference. Our method incorporates a gating mechanism and LID loss, enabling transformer experts to learn language-specific information. By combining gated transformer experts with shared transformer layers, we construct multilingual transformer blocks and utilize linear experts to effectively regularize the joint network. The curriculum training scheme leverages LID to guide the gated experts in improving their respective language performance. Experimental results on a bilingual task involving English and Spanish demonstrate significant improvements, with average relative word error reductions of 12.5% and 7.3% compared to the baseline bilingual and monolingual models, respectively. Notably, our method achieves performance comparable to the upper-bound model trained and inferred with oracle LID. Extending our approach to trilingual, quadrilingual, and pentalingual models reveals similar advantages to those observed in the bilingual models, highlighting its ease of extension to multiple languages.

## 1. INTRODUCTION

While end-to-end (E2E) models have shown rapid progress in automatic speech recognition (ASR) [1-8], there is a significant demand for multilingual ASR models, considering that more than 60% of the world's population can speak more than two languages [10]. Numerous efforts have been made to develop E2E multilingual models [11-26], which can achieve comparable or even superior ASR performance compared to monolingual baselines by incorporating language identification (LID) information through one-hot or learnable embedding vectors to differentiate between languages. However, in order to build streaming multilingual ASR systems that can perform similarly to monolingual ones in practical applications, it is crucial not to require users to input any LID information during model inference. One potential solution is to infer LID as an embedding vector and attach it to the input features [18, 19, 21]. However, this approach either leads to limited improvement due to inaccurate LID predictions or introduces additional latency for reliable LID prediction [18, 19].

In this paper, we introduce gated transformer with auxiliary LID loss and linear experts, aiming to improve multilingual speech recognition. The proposed gated transformer experts can make compact models while facilitating effective sharing of speech information across different languages. Linear experts serve to regulate the joint network output, thereby significantly improving model training stability. Additionally, we propose a curriculum training strategy that leverages LID input during training to enhance the experts' learning of language-specific information. Notably, our model eliminates the need for any LID input from users during inference. Experimental results on English and Spanish bilingual models demonstrate a 12.5% relative word error (WER) reduction compared to the bilingual model baseline without LID input, a performance similar to that achieved by the bilingual model with oracle LID input. Moreover, our bilingual model outperforms monolingual baselines. Furthermore, our methods extend successfully to trilingual, quadrilingual, and pentalingual models, with only a marginal increase in model size, achieving similar success as the bilingual models.

## 2. RELATED WORK

The concept of experts has been previously applied to ASR in [27, 28], and it has also been explored to address the challenge of bilingual code-switching, as demonstrated in [29, 30], where a dedicated encoder serves as an expert for a specific language. A gate function is utilized to combine the outputs from different experts, without the application of any LID loss to regularize their outputs. In [31], informed experts based on RNN-transducer with LID input are employed for multilingual ASR. A key distinction between our method and the approach presented in [31] is informed experts utilize Oracle LID as input for both training and inference. In contrast, our method eliminates the need for users to provide LID information during inference. While [31] briefly discusses the use of an LSTM gate to replace the LID based language gate, the training of the LSTM gate does not incorporate any LID loss, thereby making it unable to learn language-specific information. In our design, we incorporate a gating mechanism associated with LID loss and utilize curriculum training, allowing the experts to genuinely acquire language-dependent information that can effectively distinguish between languages during inference. [20] proposes a configurable multilingual model that is trained once and can be adapted to different language combinations. It utilizes linear language experts in both the encoder and prediction networks.

## 3. MULTILINGUAL TRANSFORMER TRANSDUCER WITH GATED LANGUAGE EXPERTS

### 3.1 Transformer transducer model

A neural transducer model [4] has three components: an acoustic encoder, a label prediction network, and a joint network. Neural transducer models can use different types of models as encoders such as LSTMs in RNN-T [4] and transformers [7, 8, 9, 17, 20, 21, 22] in transformer transducer (T-T). In this study, we use T-T as the backbone model for the development. Each transformer module in the encoder network is constructed from a multi-head self-attention layer followed by a feed-forward layer. The loss function of neural transducer models is the negative log posterior of output target label $y$ given input acoustic feature $x$ and is defined as

$$L_{rnnt} = -\log P(y|x) \quad (1)$$

which is calculated by forward-backward algorithm in [4].

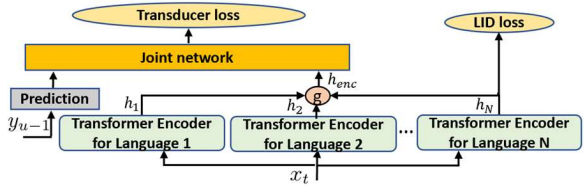

Figure 1: Architecture of multilingual T-T model with separated transformer encoders for different languages

### 3.2 Gated language transformer experts in encoder

Encoder is the most important component in T-T models. In multilingual speech recognition, if all languages share one encoder, different languages may affect the model performance since they can be confused by each other as discussed in [22]. In this work, we associate each language with its own specific transformer encoder as shown in Figure 1. Different encoders can be combined with a gate $g$ that is defined as

$$O = W_o(\tanh(\sum_{i=0}^{N} W_i h_i))) \quad (2)$$

$$g = Softmax(O) \quad (3)$$

where $h_i$ is the encoder embedding from language $i$, and $W_i$ is a linear matrix that is associated with each language, $N$ is the number of languages. Then the whole encoder network embedding output is

$$h_{enc} = \sum_{i=0}^{N} g_i h_i \quad (4)$$

Since gate $g$ combines encoder embedding from different languages, the encoder networks themselves do not realize which language they should serve as the corresponding transformer encoder. In order to make the encoder networks to learn their own corresponding languages, a LID cross entropy (CE) loss is proposed as

$$L_{lid} = CE(O) \quad (5)$$

Therefore, the overall loss is defined as a combination of the original transducer loss and the LID loss as following

$$L = L_{rnnt} + \lambda L_{lid} \quad (6)$$

where $\lambda$ is the weight to adjust the ratio of these two losses.

One drawback of the above design is each language has its own encoder that makes the model difficult scale up when the number of languages increases. Also, separated encoders for different languages may not be an optimal choice since there are still lots of acoustic conditions, speaker voice characteristics, and even pronunciation similarity that could be shared across different languages. Therefore, we propose another more effective and compact encoder structure for the T-T based multilingual ASR as shown in Figure 2. Instead of building fully separated encoders for each language, different languages can share transformer layers while each language can still have their own corresponding transformer experts that are combined with gate $g$ as defined in Equations (2) and (3) while the LID loss function is defined as following

$$L_{lid} = CE(\sum_{l=1}^{M} O_l) \quad (7)$$

where $O_l$ is the logit output as in Equation (2) but from different layer $l$. Language dependent experts, their gate, and the shared transformer layers can construct a multilingual transformer block as shown in the dotted lines of Figure 2. In a T-T multilingual model, there can be several multilingual transformer blocks included. In addition, we can also apply the shared transformer layers at the bottom of the network since the input filterbank speech features share lots of common characteristics from different languages. With this new structure, the network can share common speech information while learning linguistic knowledge from different languages. Besides, the multilingual model size can be easily controlled by the number of blocks, which is beneficial for scaling up the multilingual models to more languages.

### 3.3 Gated language linear experts for joint network

Joint network in T-T model combines both acoustic and language information from encoder and prediction networks. Inspired by [20], instead of adding language specific gated linear experts on both encoder and prediction networks, linear experts can be directly applied to the output of the joint network as shown in Figure 2. Let's define $W = \{w_1, w_2, ..., w_N\}$ is linear matrix combination of $w_i$ ($i = 1$ to N) that is a linear expert matrix corresponding to a language $i$. After multiplying this language gated linear matrix, the new joint network output is defined as

$$h'_{joint} = h_{joint} \, W \, G_L \qquad (8)$$

where $h_{joint}$ is the output of joint network and $G_L = \{g_{L_1}, g_{L_2}, ..., g_{L_N}\}$ is the gating network that is an affine LID projection referred as a "LID gate" that is controlled by the LID input. In order to get better model performance, we do leverage oracle LID in model training, but during inference, we don't need users to input any LID information, which we elaborate in more detail in section 3.4. In addition, a layer normalization is applied on $h'_{joint}$ to stabilize the training.

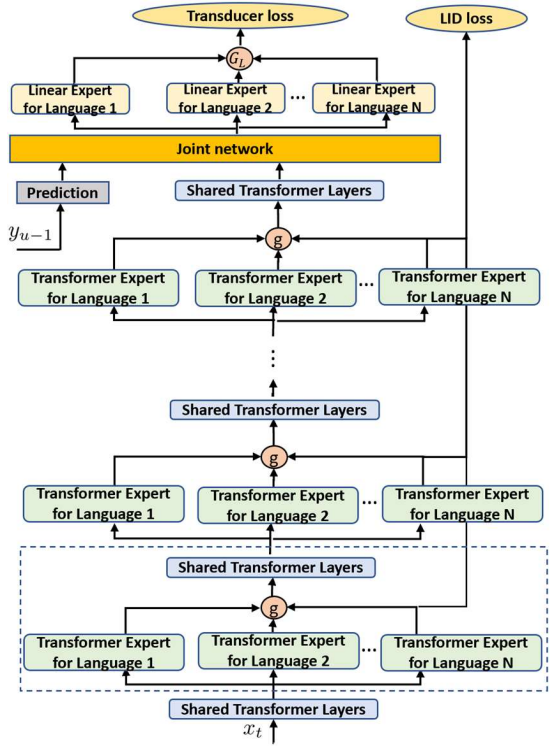

Figure 2: Architecture of multilingual T-T models with shared transformer layers at the bottom, multilingual transformer blocks (one block is defined as in dotted lines) including gated language transformer experts and shared layers, and linear experts for joint network

### 3.4 LID-based curriculum training strategy

Even though the LID loss is applied to enforce transformer experts to learn the corresponding speech information for its own language, the network can still be confused by the languages with the similar word pronunciations and the same writing letters, especially at the early stage of training which could lead to the model performance degradation. In [29], a seed model with explicitly leveraging LID information is pre-trained to relieve this language confusion issue. In our method, a single LID is passed to the LID gate to set its corresponding values to 1 as mentioned in section 3.3 at the early stage of model training to guide the transformer experts to learn their own languages. When training is going on, we also pass multilingual LIDs from all languages to the LID gate while still keeping a portion of passing one single LID to have the gradual transition from one LID training to multiple LID training, which is called as the curriculum training strategy for LID input. At the final stage of model training, only multiple LIDs for all languages are passed to the LID gate. Let's take the bilingual model as an example. At the beginning of training, we only pass LID vectors [0, 1] or [1, 0] to the LID gate for different input languages, and then in the middle stage of training, we pass 1hot vectors [0, 1], [1,0], or 2-hot vector [1,1] to the model training with a probability of p for 1 hot vectors, and 1- p for 2-hot vector. p decreases when training goes on. At the final stage of training, p reduces to 0, and we only pass 2-hot vector [1,1] in model training. Then the Equation (2) is further improved as

$$O_l = W_o(\tanh(\sum_{i=0}^{N} W_{l_i} \, h_{l_i} \, g_{L_i}))) \qquad (9)$$

where $l$ is the layer number of multilingual transformer block. Only one $g_{L_i}$ corresponding to the language $i$ in the language gate $G_L$ is set to 1 at the beginning of training, and all values in $G_L$ are set to 1 for the final stage of training (Note: we omit to draw $G_L$ for the transformer experts in Figure 2 to make the figure less complicated). During inference, the multi-hot LID vector with all its element value 1 is passed to the LID gate by the system for multilingual speech recognition and there is no need for users to input any specific LID information. This training strategy also applies to linear experts of joint network described in Equation (8) in model training.

## 4. EXPERIMENTAL SETUPS

### 4.1 Language and data

We develop our multilingual T-T models to support up to five languages which are English (EN), Spanish (ES), German (DE), Italian (IT), and French (FR). For all these languages, both training and test data are transcribed and anonymized with personally identifiable information removed. Test data includes both in-domain data sampled from the same distribution as training, and also out-of-domain data that is different from training. The training and test data amount per language is summarized in Table 1.

## 4.2 Model structures and training configurations

In our baseline T-T models, 18 basic transformer modules with 320 hidden nodes, 8 attention heads, and 2048 feedforward nodes are used as the encoder; 2 LSTM layers with 1024-dimensional embedding and hidden layer are used in the prediction network. The basic transformer modules are also applied as the transformer experts in multilingual transformer blocks as shown in Figure 2 without any structure change. 80-dimensional log-Mel filterbank are used with 25 milliseconds (ms) windows and 10ms shift. LID vectors are appended to input features in both model training and inference. Two convolutional layers are applied to get features with 40ms sampling rate. The input acoustic feature sequence is segmented into chunks with a chunk size of 4 and chunks are not overlapped. In addition, we also apply 18 left chunks to leverage history acoustic information. An effective mask strategy to truncate history and allow limited future lookahead information has been designed as in [9]. The learning rate warmup strategy is the same as in [32]. Each training mini-batch consists of utterances from all languages, sampled according to their training data distributions. We train BPE models to generate token lists for each language separately, and then merge token lists together as the multilingual model output. For monolingual, bilingual, trilingual, quadrilingual, and pentalingual models, their output tokens are 4k, 7k, 10k, 12k, and 14k, respectively.

Table 1: Train and test data per language (in hours)

| Language | Train | Test |
|---|---|---|
| EN | 23,035 | 208 |
| ES | 3,770 | 33 |
| DE | 2,893 | 38 |
| IT | 3,345 | 19 |
| FR | 3,176 | 33 |
| Total | 36,219 | 331 |

## 5. RESULTS

### 5.1 English and Spanish bilingual model

We started investigating our proposed methods in Section 3 using English and Spanish bilingual models. The baseline English and Spanish bilingual model, referred to as B2 in Table 2, was trained by pooling all the data from both languages without providing LID information to the model. The model size for B2 is 80M. Additionally, we trained monolingual models as another baseline. Table 2 also provides information about the parameter sizes for different model structures.

From Table 2, it can be observed that B2 got an average WER of 16.0%, which is higher than the average WER of 15.1% obtained by the monolingual models. Furthermore, we trained a bilingual model, B1, with the oracle LID provided as input during both training and inference stages to establish an upper bound for bilingual model performance. B1 obtained a relative WER reduction of 13.1% compared to B2. The oracle LID was represented as a one-hot vector appended to the input features, following a similar approach as in [16].

Table 2: WERs (%) and parameter numbers (M) for English and Spanish bilingual models

| Model | Params | EN | ES | Avg |
|---|---|---|---|---|
| Monolingual | 78*2 | 13.2 | 16.2 | 15.1 |
| B1 Oracle LID | 80 | 12.8 | 14.9 | 13.9 |
| B2 baseline without LID | 80 | 14.9 | 17.1 | 16.0 |
| B3 fully seperated encoder | 133 | 13.6 | 16.0 | 14.8 |
| B4 6 transformer blocks | 100 | 13.1 | 15.5 | 14.3 |
| B5 3 transformer blocks | 90 | 13.2 | 15.8 | 14.5 |
| B6  + joint linear expert | 90.5 | 13.1 | 15.6 | 14.4 |
| B7    + CT for LID input | 90.5 | 13.0 | 14.9 | 14.0 |

We then trained a bilingual model, B3, with fully separated encoders as shown in Figure 1. To prevent model divergence, the gating mechanism was based on the last 3$^{rd}$ layer embeddings from both encoders, with the last two layers being used as shared layers. B3 achieved a WER of 14.8%, representing a 7.5% relative WER reduction compared to B2. However, the encoder's model size almost doubled, and the overall parameter count increased to 133M, making it less feasible for extension to additional languages.

As proposed in Section 3.2, we introduced model B4, which was trained with 6 multilingual transformer blocks (illustrated by dotted lines in Figure 2) to mitigate the model size expansion. Each multilingual transformer block consisted of two shared layers, and no shared layers were applied at the bottom layer before the multilingual transformer blocks. B4 not only had a more compact structure with 100M model parameters but also encouraged the sharing of speech and language information among different languages. B4 achieved relative WER reductions of 10.6% and 3.4% compared to B2 and B3, respectively. To further reduce the model size, we decreased the number of multilingual transformer blocks from 6 to 3 and added 9 shared transformer layers before the multilingual transformer blocks to train model B5. This resulted in a model that was only 1.4% worse in terms of relative WER compared to B4, but with 10M fewer model parameters. We experimented with adding transformer blocks at different locations, and placing them near the model output proved to be the most effective for B5.

Based on B5, we proceeded to train the B6 model using joint linear experts as proposed in Section 3.3. This led to a slight improvement in performance, reducing the average WER from 14.5% to 14.4% with a mere increase of 0.5M model parameters. Furthermore, the training recipe for B6 provided much greater stability, particularly when incorporating more languages into the multilingual model building process.

Finally, we applied the curriculum strategy (CT) for LID input to guide the language-dependent transformer and linear experts in learning their respective languages. This resulted in model B7, which achieved relative WER reductions of 12.5% and 7.3% compared to the B2 baseline and monolingual models, respectively. B7 even achieved a similar average result as the upper model of B1 (14.0% vs. 13.9%). Additionally, we increased the number of parameters for B2 to match B7, and the slight parameter increase did not significantly impact the WER.

Table 4: WERs (%) and parameter numbers (M) for trilingual, quadrilingual, and pentalingual models

| Languages | Params | EN | ES | DE | IT | FR | Avg |
|---|---|---|---|---|---|---|---|
| Monolingual | 78*n (n=3,4,5) | 13.2 | 16.2 | 15.7 | 13.2 | 16.5 | 15.1 |
| T1 Trilingual Oracle LID | 82 | 12.8 | 14.5 | 15.5 | - | - | 14.3 |
| T2 Trilingual without LID | 82 | 14.9 | 17.2 | 16.0 | - | - | 16.0 |
| T3 Gated Expert Trilingual | 100.5 | 12.9 | 14.6 | 15.4 | - | - | 14.3 |
| Q1 Quadrilingual Oracle LID | 84 | 12.9 | 14.6 | 15.2 | 12.0 | - | 13.7 |
| Q2 Quadrilingual without LID | 84 | 15.0 | 17.6 | 16.2 | 14.7 | - | 15.9 |
| Q3 Gated Expert Quadrilingual | 110.5 | 13.0 | 14.8 | 15.4 | 12.2 | - | 13.9 |
| P1 Pentalingual Oracle LID | 86 | 12.9 | 14.5 | 15.3 | 12.0 | 15.2 | 14.0 |
| P2 Pentalingual without LID | 86 | 15.2 | 18.2 | 16.5 | 15.6 | 16.6 | 16.4 |
| P3 Gated Expert Pentalingual | 120.5 | 13.2 | 14.8 | 15.5 | 12.1 | 15.5 | 14.2 |

We also evaluated the performance of the models, B2 and B7, on a Spanish/English code-switching test set, where English words were included as entity names. The results in Table 3 demonstrate that B7 achieved a significant relative WER reduction of 9.3% compared to the B2 baseline. This further reinforces the effectiveness of our methods for multilingual Automatic ASR.

Table 3: Spanish/English code-switching test set results

| Model | WER |
|---|---|
| B2 | 28.1% |
| B7 | 25.5% |

### 5.2 Extension to more languages

Building upon the success of developing English and Spanish bilingual models, we expanded our methods to construct multilingual models that encompass additional languages such as German, Italian, and French. The structure of these models closely resembles that of the bilingual model B7, but with the inclusion of dedicated transformers and linear experts for each new language. Specifically, for each added language, three language-dependent transformer and one linear experts were applied within the transformer blocks of model B7, resulting in a 10M increase in model parameters, as shown in Table 4. To measure the performance of the multilingual models, we provided monolingual models as reference points. The baseline model results, without utilizing any LID information, as well as the upper bound results from models trained and inferred with oracle LID, were presented in Table 4 for comparison with our proposed models.

When incorporating additional languages, the parameter size of the baseline models also slightly increased due to the inclusion of more token labels in the output layer. Similarly, our models with gated experts consistently demonstrated significant improvements. They achieved average WER reductions of 10.6%, 12.6%, and 13.4% over the corresponding trilingual, quadrilingual, and pentalingual baseline models without LID, respectively. Comparing our models to the upper bound models with oracle LID, we found that our models achieved similar WERs: 14.3% vs. 14.3% for trilingual models, 13.7% vs. 13.9% for quadrilingual models, and 14.0% vs. 14.2% for pentalingual models.

When examining individual languages, we observed that the WERs for Spanish, without LID, worsened as more languages were added: 17.2%, 17.6%, and 18.2% for trilingual, quadrilingual, and pentalingual models, respectively. This pattern was also observed for German and Italian. However, our models did not get the performance degradation when incorporating more languages into the multilingual models. They exhibited a consistent performance trend similar to the models trained and inferred with oracle LID information. Finally, our gated expert-based trilingual, quadrilingual, and pentalingual models consistently achieved similar or better WERs compared to the monolingual models.

### 6. CONCLUSIONS

In this paper, we present a novel approach that enhances the performance of the multilingual T-T model by leveraging gated language experts and auxiliary LID loss, without requiring any LID input from users during model inference. Our approach involves constructing multilingual transformer blocks, comprising gated transformer experts and shared layers in encoders. This framework facilitates the sharing of common speech and acoustic information through the shared layers, while the transformer experts specialize in learning language-dependent knowledge. In addition, we incorporate linear experts into the joint network output to effectively regularize joint speech acoustic and token label information, significantly improving training stability. To further enhance the learning of language-specific experts, we introduce a curriculum training strategy that incorporates LID input. Experimental results on English and Spanish bilingual models demonstrate relative WER reductions of 12.5% and 7.3%, respectively, compared to the bilingual model baseline without LID information and monolingual models. Moreover, our method achieves model performance similar to that of models trained and inferred with oracle LID. Extending our approach to five languages yields similar patterns as observed in the bilingual models.